\documentclass[a4paper]{article}

\usepackage{INTERSPEECH2019}
\usepackage{multirow}
\usepackage{pgfplots, pgfplotstable}
\usepackage{enumitem}

\pgfplotsset{compat=1.14}
\usetikzlibrary{matrix}

\title{Sequence-to-Sequence Speech Recognition with \\ Time-Depth Separable Convolutions}
\name{Awni Hannun, Ann Lee, Qiantong Xu, Ronan Collobert}

\address{Facebook AI Research}
\email{awni@fb.com, annl@fb.com, qiantong@fb.com, locronan@fb.com}

\begin{document}

\maketitle

\begin{abstract}
  We propose a fully convolutional sequence-to-sequence encoder architecture with a simple and efficient decoder. Our model improves WER on LibriSpeech while being an order of magnitude more efficient than a strong RNN baseline. Key to our approach is a time-depth separable convolution block which dramatically reduces the number of parameters in the model while keeping the receptive field large. We also give a stable and efficient beam search inference procedure which allows us to effectively integrate a language model. Coupled with a convolutional language model, our time-depth separable convolution architecture improves by more than 22\% relative WER over the best previously reported sequence-to-sequence results on the noisy LibriSpeech test set.
\end{abstract}
\noindent\textbf{Index Terms}: speech recognition, sequence-to-sequence, neural networks

\section{Introduction}

Sequence-to-sequence models with attention have been used for speech recognition~\cite{chorowski2015} since their inception in machine translation~\cite{bahdanau2014neural, cho2014learning, sutskever2014sequence}. These models have yielded state-of-the-art results in some settings~\cite{chiu2018state}, however; approaches such as CRF style end-to-end models~\cite{collobert2016wav2letter, hannun2014deep} and more traditional HMM based models~\cite{povey2011kaldi} are often superior.

While sequence-to-sequence models sometimes generalize well in speech recognition, they often come with a big hit to efficiency. The encoder typically consists of several layers of large bidirectional LSTMs~\cite{chan2016, zhang2017very}. The decoder also uses a number of inefficient and sequential techniques. Efficiency is useful for fast training and evaluation times and is critical to the massive scale used in the semi-supervised and weakly supervised regimes~\cite{edunov2018understanding, mahajan2018exploring}. 

In this work we develop a highly efficient sequence-to-sequence model which gives state-of-the-art results for non speaker adapted models on both LibriSpeech test sets~\cite{panayotov2015librispeech}. Key to our approach is a \emph{fully convolutional} encoder with a time-depth separable (TDS) block structure. Our TDS convolution improves in WER over an RNN baseline and due to the parallel nature of the computation is much more efficient. We also discard slow and sequential techniques previously thought to be important to the accuracy of these models. These include neural content attention, location based attention, and scheduled sampling. In turn, we give more efficient alternatives.

Also key to our approach is a highly efficient and stable beam search inference procedure. Unlike previous work~\cite{chorowski2016towards}, accuracy does not degrade with very large beam sizes. This enables us to better leverage the constraint of a convolutional language model which gives substantial improvements in WER over a simple n-gram baseline.

\section{Model}
\label{sec:model}

\begin{figure}
    \centering
    \includegraphics[width=\linewidth]{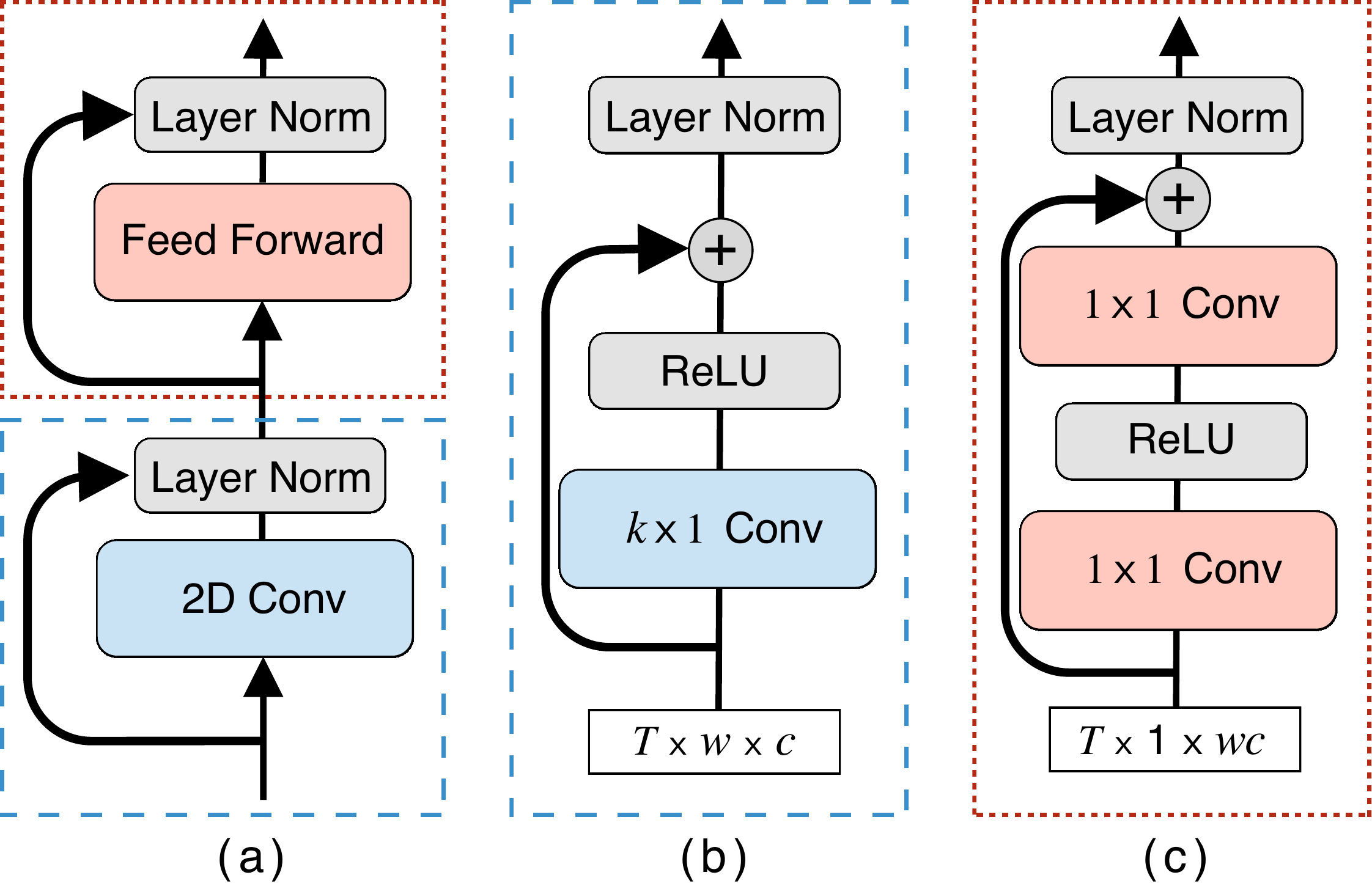}
    \caption{The TDS convolution model architecture. (a) The sub-blocks of the TDS convolution layer are (b) a 2D convolution over time followed by (c) a fully connected block.}
    \label{fig:model_arch}
\end{figure}

We consider an input utterance $X = [X_1, \ldots, X_T]$ and an output transcription $Y = [y_1, \ldots, y_U]$. The sequence-to-sequence model \emph{encodes} $X$ into a hidden representation and then \emph{decodes} the hidden representation into a sequence of predictions for each output token. The encoder is given by 
\begin{equation}
    \begin{bmatrix} K\\ V \end{bmatrix} = \text{encode}(X)
\end{equation}
where $K = [K_1, \ldots, K_T]$ are the keys and $V = [V_1 \ldots, V_T]$ are the values. The decoder is given by 
\begin{align}
    Q_u &= g(y_{u-1}, Q_{u-1}) \label{eq:attend} \\
    S_u &= \text{attend}(Q_{u}, K, V) \\
    P(y_u \mid X, y_{< u}) &= h(S_u, Q_u)
\end{align}
Here $g(\cdot)$ is an RNN which encodes the previous token and query vector $Q_{u-1}$ to produce the next query vector. The attention mechanism $\text{attend}(\cdot)$ produces a summary vector $S_u$, and $h(\cdot)$ computes a distribution over the output tokens.

\subsection{Time-Depth Separable Convolutions}
\label{sec:tdsconv}

Our proposed time-depth separable (TDS) convolution block (see Figure~\ref{fig:model_arch}) partially decouples the aggregation over time from the mixing over channels. This allows us to increase the receptive field of the model with a negligible increase in the number of parameters. In preliminary experiments we find that the TDS convolution block generalizes much better than other deep convolutional architectures~\cite{collobert2016wav2letter, gehring2017convolutional} and needs fewer parameters. Another benefit of our block structure is it can be implemented efficiently using a standard 2D convolution.

The block starts with a layer of 2D convolution which operates over an input of shape $T \times w \times c$ and produces an output of shape $T \times w \times c$ where $T$ is the number of time-steps, $w$ is the input width and $c$ is the number of input (and output) channels. The kernels are size $k \times 1$. The total number of parameters in this layer is $kc^2$ which can be made small by keeping $c$ small. We follow the convolution with a ReLU non-linearity.

We then view the output of the convolution as $T \times 1 \times wc$ and apply a fully-connected layer, which is a sequence of two $1 \times 1$ convolutions (i.e. linear layers) with a ReLU non-linearity in between. We add residual connections~\cite{zhang2017very, he2016deep} and layer normalization~\cite{ba2016layer} after the convolution and the fully connected layer. The layer normalization is over all dimensions for a given example including time.

The TDS architecture has three sub-sampling layers each with a stride of 2 for a total sub-sampling factor of 8. We also increase the the number of output channels at each sub-sampling layer since we compress the information in time. For simplicity these layers do not have residual connections and are only followed by a ReLU and layer normalization.

\subsection{Efficient Decoder}
\label{sec:efficientdecoder}

The decoder is sequential in nature since to compute the next output requires the previous prediction. However, at training time we use teacher forcing--the previous ground truth is used in place of the previous prediction. In principle, this allows us to compute all output frames simultaneously. The outputs of the RNN given by $g(\cdot)$ cannot be computed in parallel, however; unrolling the computation and making a single call to an efficient CuDNN~\cite{chetlur2014cudnn} implementation is much faster than calling $U$ separate kernels. After the following optimizations, the decoder accounts for less than 10\% of the total iteration time.

Techniques such as scheduled sampling~\cite{bengio2015scheduled}, input feeding~\cite{luong2015effective} and location-based attention~\cite{chorowski2015} introduce a sequential dependency in the decoder. We discard these techniques in favor of approaches which can be computed in parallel. We simply do not use input feeding and location-based attention as we find that we can achieve good WERs without them. We replace scheduled sampling with random sampling (section~\ref{sec:randomsampling}).

We use an inner-product key-value attention which can be implemented much more efficiently than a neural attention. For a single example the attention is given by 
\begin{align}
    S = V \cdot \text{softmax}\left(\frac{1}{\sqrt{d}} K^\top Q \right)
\end{align}
We scale the inner products by the inverse square root of their hidden dimension $d$. This improves convergence and helps the model learn an alignment. However, we do not see a consistent improvement in generalization~\cite{vaswani2017attention}.

\subsubsection{Random Sampling}
\label{sec:randomsampling}

Scheduled sampling~\cite{bengio2015scheduled} limits exposure bias by bringing the training conditions closer to the testing conditions. However, it introduces a sequential dependency in the decoder, since it sometimes uses the previous prediction at the next time-step.

Instead, we propose random sampling, where the previous prediction is replaced with a randomly sampled token~\cite{wang2018switchout}. First we decide with probability $P_{\text{rs}}$ to sample a given input token. If we sample, then choose a new token from a uniform distribution. This allows us to vectorize the implementation as follows:
\begin{enumerate}
    \item Sample $U$ random numbers $c_j$ uniformly from $[0, 1]$.
    \item Set $R = [r_1, \ldots, r_U]$ where $r_j = \mathbb{I}(c_j > P_{\text{rs}})$ and $P_{\text{rs}}$ is the sampling probability.
    \item Sample a vector $Z$ of $U$ tokens. We use a uniform distribution over the output tokens not including end-of-sentence (EOS).
    \item Construct $\hat{Y} = R \circ Z + (1-R) \circ Y$.
\end{enumerate}
As we show later, random sampling improves WER.

\subsection{Soft Window Pre-training}

We propose a simple soft attention window pre-training scheme to enable the training of very deep convolutional encoders. Compared to prior work~\cite{zeyer2018}, our approach is simple to implement, results in negligible additional computational expense, and needs very little tuning.

We encourage the model to align the output at uniform intervals along the input by penalizing attention values which are too far from the desired locations. Let $W$ be a $T \times U$ matrix with entries $W_{ij} = (i - \frac{T}{U} j)^2$. The matrix $W$ encodes the (squared) distance between the $i$-th input and the $j$-th output assuming the outputs are spaced at uniform intervals along the input -- hence the scaling factor $T / U$. We apply $W$ to the attention as follows

\begin{equation}
    S = V \cdot \text{softmax}\left(\frac{1}{\sqrt{d}} K^\top Q 
                    - \frac{1}{2 \sigma^2}W \right)
\end{equation}
The term $\sigma$ is a hyper-parameter which dampens the effect of $W$. The application of $W$ is equivalent to multiplying the normalized attention vector (i.e. after the softmax) by a Gaussian shaped mask. In that respect, $\sigma$ is simply the standard deviation of the Gaussian.

We use the window pre-training for the first few epochs and then switch it off. This is sufficient to enable the model to learn an alignment and converge. In general $\sigma$ does not need to be tuned when model hyper-parameters change. An exception is when the amount of sub-sampling in the encoder changes, $\sigma$ should change accordingly.

\subsection{Regularization}

We use three additional forms of regularization to control overfitting and improve the generalization of the model.

\subsubsection{Dropout}

First we apply dropout~\cite{hinton2012improving} after each layer in each block of the encoder. We apply dropout after the non-linearity and prior to layer normalization. We do not use any dropout in the decoder.

\subsubsection{Label Smoothing}

We use label smoothing~\cite{szegedy2016rethinking} to reduce over-confidence in predictions. As in machine translation~\cite{vaswani2017attention}, we find that label smoothing hurts loss on the dev set but improves WER.

\subsubsection{Word Piece Sampling}

We use word pieces~\cite{sennrich2015neural} as outputs following the Unigram Language Model approach~\cite{kudo2018subword}. During training, we sample word piece representations for a given transcription~\cite{kudo2018subword}, but unlike prior work, we sample at the word-level instead of the sentence-level. For each word, with probability $1 - P_{\text{wp}}$ we take the most likely word piece representation or with probability $P_{\text{wp}}$ uniformly sample over the top-ten most likely alternatives.

\section{Beam Search Decoding}

We use an \textit{open-vocabulary} beam search decoder which optimizes the following objective

\begin{equation}
    \log P_{\text{s2s}}(Y \mid X) + \alpha \log P_{\text{LM}}(Y) + \beta |Y|
\end{equation}
The term $|Y|$ counts the number of tokens in $Y$. In the above, $\alpha$ is the LM weight and $\beta$ is the token insertion term.

\subsection{Stabilizing Beam Search}

Sequence-to-sequence beam search decoders are known to be unstable sometimes exhibiting worse performance with an increasing beam size~\cite{chorowski2016towards}. We use two techniques to stabilize the beam search. This allows our model to extract more value from the integration of an LM, since we can use a large beam size to effectively search over the space of possible hypotheses.

\subsubsection{Hard Attention Limit}

We do not allow the beam search to propose any hypotheses which attend more than $t_{\text{max}}$ frames away from the previous attention peak. In practice we find that $t_{\text{max}}$ only needs to be tuned once for a given data set and can otherwise remain unchanged.

\subsubsection{End-of-sentence Threshold}

In order to bias the search away from short transcriptions, we only consider end-of-sentence (EOS) proposals when the score is greater than a specified factor of the best candidate score
\begin{align}
    \label{eq:eosthresh}
    \log P_u(\text{EOS} \mid y_{<u}) > \gamma \cdot \max_c \log P_u(c \mid y_{<u})
\end{align}
Like the hard attention limit, we find the parameter $\gamma$ only needs to be tuned once for a given data set.

\subsection{Efficiency}
\label{sec:beamsearchefficiency}

We use a few heuristics to further improve the efficiency of the beam search. First, we set a beam threshold~\cite{collobert2016wav2letter} to prune hypotheses in the beam which are below a fixed range from the best hypothesis so far.

We also apply a threshold when proposing new candidate tokens to the current set of hypotheses in the beam. Similar to Equation~\ref{eq:eosthresh}, we require that the proposed token score satisfy
\begin{align}
    \log P_u(y \mid y_{<u}) > \max_c \log P_u(c \mid y_{<u}) - \eta
\end{align}

Finally, we batch compute the updated set of probabilities for every candidate in the beam, so only one forward pass is required at each step. These techniques result in a fast decoding time even with a deep convolutional LM and a large beam.

\section{Experiments}
\label{sec:experiments}

We perform experiments on the full 960-hour LibriSpeech corpus~\cite{panayotov2015librispeech}. Our best encoder has two 10-channel, three 14-channel and six 18-channel TDS blocks. We use three 1D convolutions to sub-sample over time, one as the first layer and one in between each group of TDS blocks. Kernel sizes are all 21$\times$1. A final linear layer produces the 1024-dimensional encoder output. The decoder is a one-layer GRU with 512 hidden units. Weights are initialized from a uniform distribution $\mathcal{U}(-\sqrt{4/f_{in}}, \sqrt{4/f_{in}})$, where $f_{in}$ is the fan-in to each unit. 

Input features are 80-dimensional mel-scale filter banks computed every 10-ms with a 25-ms window. We use 10k word pieces computed from the \textit{SentencePiece} toolkit~\cite{kudo2018sentencepiece} as the output token set. All models are trained on 8 V100 GPUs with a batch size of 16 per GPU. We use synchronous SGD with a learning rate of 0.05, decayed by a factor of 0.5 every 40 epochs. We clip the gradient norm to 15. The model is pre-trained for three epochs with the soft window and $\sigma=4$. We use 20\% dropout, 5\% label smoothing, 1\% random sampling and 1\% word piece sampling.

We train two word piece LMs on the 800M-word text-only data set. The first is a 4-gram trained with KenLM~\cite{heafield2011kenlm} and the second is a convolutional LM (ConvLM)~\cite{dauphin2017language} using the same model architecture and training strategy as \cite{zeghidour2018fully}. We use a beam size of 80, set $t_{\max}=30$, the EOS penalty $\gamma = 1.5$ and $\eta = 10$. The LM weight and token insertion terms are cross-validated with each dev set and LM combination. We use the {\it wav2letter++} framework to train and evaluate our models~\cite{pratap2018wav2letter++}.


\begin{table}
    \centering
    \setlength{\tabcolsep}{3pt}
    \caption{A comparison of the TDS conv model to other models on the Librispeech Dev and Test sets.}
    \label{tab:wers}
    \begin{tabular}{l c c c c} 
        \toprule
         \multirow{2}{*}{Model} & \multicolumn{2}{c}{Dev WER} & \multicolumn{2}{c}{Test WER} \\
               & clean & other & clean & other \\
      \midrule
        \multicolumn{3}{l}{{\it hybrid, speaker adapted}} & & \\
        CAPIO (single) \cite{han2017capio} + RNN & 3.12 & 8.28 & 3.51 & 8.58 \\ 
        CAPIO (ensemble) \cite{han2017capio} + RNN & 2.68 & 7.56 & 3.19 & 7.64 \\
       \midrule
        CNN ASG \cite{zeghidour2018fully} + ConvLM & 3.16 & 10.05 & 3.44 & 11.24 \\
        RNN S2S \cite{zeyer2018} & 4.87 & 14.37 & 4.87 & 15.39 \\
        RNN S2S \cite{zeyer2018} + 4-gram & 4.79 & 14.31 & 4.82 & 15.30  \\
        RNN S2S \cite{zeyer2018} + LSTM   & 3.54 & 11.52 & 3.82 & 12.76 \\
       \midrule
       \midrule
        {\bf TDS conv} & 5.04 & 14.45 & 5.36 & 15.64 \\
        {\bf TDS conv} + 4-gram & 3.75 & 10.70 & 4.21 & 11.87 \\
        {\bf TDS conv} + ConvLM & 3.01 & 8.86 & 3.28 & 9.84 \\
       \bottomrule
    \end{tabular}
\end{table}

\subsection{Results}

Table~\ref{tab:wers} compares the TDS model with three other systems. The CAPIO system is a hybrid HMM-DNN with speaker adaptation~\cite{han2017capio}. The other two are end-to-end models, one using the CRF-style ASG loss~\cite{zeghidour2018fully} and the other a sequence-to-sequence model with an RNN encoder~\cite{zeyer2018}.

Our proposed model achieves a state-of-the-art for end-to-end systems of 3.28 WER on test clean and 9.84 WER on test other. Compared with the RNN-based encoder~\cite{zeyer2018}, the TDS model improves WER by 14.1\% on test clean and 22.9\% on test other with nearly a factor of 4 reduction in parameters (136M vs. 37M). The TDS model benefits more from an external LM. This could be due to (1) a better loss on the correct transcription and (2) a more effective beam search.

\subsection{Model Variations}
Table~\ref{tab:sensitivity} shows results from varying the number of TDS blocks, the number of parameters, the word piece sampling probability and the amount of random sampling. For each setting we train three models and report the best and the average WER.  

We reduce the number of parameters without changing the receptive field by reducing the number of channels in each group of TDS blocks from (10, 14, 18) to (10, 12, 14) or (10, 10, 10). The model is very sensitive to decreasing the number of parameters. We also examine the effect of varying the number of TDS blocks without changing the number of parameters or the receptive field. For 9 TDS blocks we use (14, 16, 20) channels with $k=27$, and for 12 TDS blocks we use (10, 16, 16) channels with $k=19$. We show that a small amount of word piece sampling is helpful. With a higher $P_{wp}$ the model sometimes converges poorly, likely due to the variability in the targets. A small amount of random sampling is also helpful. Finally, when we remove soft window pre-training, the model takes much longer to converge and achieves a worse result. The soft window clearly helps guide the attention early in training.

\begin{table}
    \centering
    \caption{The sensitivity of our model to architecture and regularization hyper-parameters. The parameter $N$ is the number of TDS blocks, $P_{\text{wp}}$ is the word piece sampling rate, and $P_{\text{rs}}$ is the random sampling rate. Missing entries correspond to the value in the first row. We report the lowest WER over three runs along with the mean in parentheses using a beam size of 1 and no LM.}
    \label{tab:sensitivity}
    \begin{tabular}{c c c c | c c} 
    \toprule
    \multirow{2}{*}{$N$} & params & \multirow{2}{*}{$P_\text{wp}$} & \multirow{2}{*}{$P_\text{rs}$} & Dev & Dev \\
        & ($\times 10^6$)  & & & Clean & Other \\
    \midrule 
        11 & 36.5 & 1\% & 1\% & 5.04 (5.13) & 14.45 (14.77) \\
    \midrule
           & 24.4 &     &      & 5.36 (5.45) & 15.16 (15.24) \\
           & 14.9 &     &      & 5.95 (5.99) & 16.25 (16.44) \\
        9  &     &     &      & 5.18 (5.27) & 15.34 (15.37) \\
        12 &     &     &      & 5.10 (5.33) & 14.99 (15.26) \\
           &     & 0\% &      & 5.25 (5.32) & 14.89 (15.00) \\
           &     & 2\% &      & 5.04 (5.46) & 14.88 (15.41)  \\
           &     &     & 0\%  & 5.08 (5.24) & 15.00 (15.21) \\
           &     &     & 5\%  & 5.11 (5.25) & 14.65 (14.80) \\
    \midrule
    \multicolumn{4}{c |}{No soft window pre-training} & 5.55 (5.58) & 14.99 (15.30) \\
    \bottomrule
    \end{tabular}
\end{table}

\begin{figure}
\begin{minipage}{.5\linewidth}
  \begin{tikzpicture}
\begin{axis}[
  title=({\bf a}) dev clean,
  xlabel=Receptive Field (s),
  ylabel=WER,
  ymax=8,
  xtick=data,
  legend style={at={(0.62, .78)},anchor=south west},
  width=1.2\columnwidth,
  height=4.5cm
]
\addplot table [y=cleanwer, x=receptivefield]{receptive_fields.dat};
\end{axis}
\end{tikzpicture}
\end{minipage}%
\hspace{2mm}
\begin{minipage}{.45\linewidth}
  \begin{tikzpicture}
\begin{axis}[
  title=({\bf b}) dev other,
  xlabel=Receptive Field (s),
  ymax=20,
  xtick=data,
  width=1.3\columnwidth,
  height=4.5cm
]
\addplot table [y=otherwer, x=receptivefield]{receptive_fields.dat};
\end{axis}
\end{tikzpicture}
\end{minipage}
\vspace{-0.3cm}
\caption{The WER as a function of the receptive field. We vary the kernel size, $k \in \{5, 9, 13, 17, 21\}$, otherwise every model has  $\sim$36.5 million parameters. We report the mean WER over three runs using a beam size of 1 and no LM.}
\label{fig:receptive_field}
\end{figure}
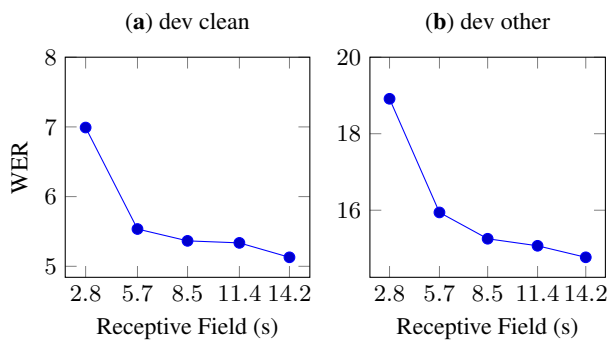

Figure~\ref{fig:receptive_field} shows the effect of the receptive field on WER. There is a sharp increase in WER when the size of the receptive field drops below a threshold. Qualitative analysis shows that the high WER is often due to catastrophic errors such as looping and skipping, a common problem for sequence-to-sequence models~\cite{chorowski2016towards}. We hypothesize that without a large receptive field, the encoder keys do not have enough context to disambiguate queries from the decoder.

\begin{figure}
\begin{minipage}{.5\linewidth}
  \begin{tikzpicture}
\begin{axis}[
  title=({\bf a}) dev clean,
  xlabel=Beam Size (log),
  ylabel=WER,
  ymax=5,
  xtick=data,
  xmode=log,  log ticks with fixed point,
  width=1.2\columnwidth,
  height=4.5cm
]
\addplot table [y=ngramclean, x=beamsize]{beam_search.dat};
\addplot table [y=convclean, x=beamsize]{beam_search.dat};
\addlegendentry{4-gram}
\addlegendentry{ConvLM}
\end{axis}
\end{tikzpicture}
\end{minipage}%
\hspace{2mm}
\begin{minipage}{.45\linewidth}
  \begin{tikzpicture}
\begin{axis}[
  title=({\bf b}) dev other,
  xlabel=Beam Size (log),
  xmode=log,  log ticks with fixed point,
  ymax=13,
  ytick={9,10,11,12,13},
  xtick=data,
  width=1.3\columnwidth,
  height=4.5cm
]
\addplot table [y=ngramother, x=beamsize]{beam_search.dat};
\addplot table [y=convother, x=beamsize]{beam_search.dat};
\addlegendentry{4-gram}
\addlegendentry{ConvLM}
\end{axis}
\end{tikzpicture}
\end{minipage}
\vspace{-0.3cm}
\caption{The WER as a function of beam size for both the 4-gram and the convLM.}
\vspace{-3mm}
\label{fig:beamsearch}
\end{figure}
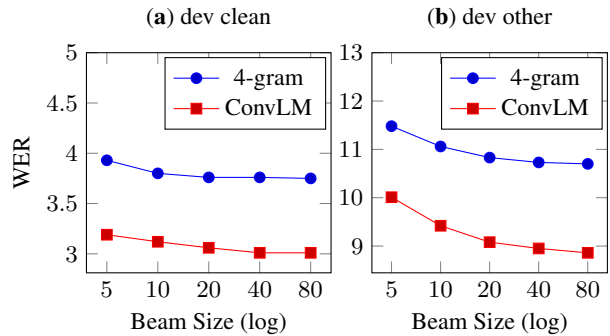

Figure~\ref{fig:beamsearch} shows how WER changes with the size of the beam. While most of the gain from including an external LM comes even at small beam size, we see consistent improvements up to a beam size of 80, particularly on dev other.

\subsection{Efficiency}

We compare the TDS conv model to a strong RNN baseline in terms of training efficiency on LibriSpeech~\cite{zeyer2018}. The RNN baseline encoder consists of six bidirectional LSTMs. Both models have a total sub-sampling factor of 8. Our best TDS architecture can complete one full epoch over the LibriSpeech training set in 7 minutes. This is more than 10$\times$ faster than our implementation of the RNN baseline and more than 4$\times$ faster than the RNN baseline encoder but with the efficient decoder described in Section~\ref{sec:efficientdecoder}.

Our beam search runs at an average rate of 0.57 and 0.93 seconds-per-sample on dev clean and other with the 4-gram LM and a beam size of 80. With the ConvLM, times increase to 0.73 and 1.20 seconds-per-sample at the same beam size.

\section{Related Work}

Our work builds on a large body of work aimed at improving sequence-to-sequence models with attention for both speech recognition~\cite{chiu2018state, chorowski2016towards, zeyer2018} and other application domains. Fully convolutional encoders have worked well in machine translation~\cite{gehring2017convolutional}. They have also given state-of-the-art results in speech recognition~\cite{zeghidour2018fully} with more structured loss functions like the AutoSegCriterion~\cite{collobert2016wav2letter}. However, we are not aware of any competitive results with fully convolutional encoders in sequence-to-sequence models for speech recognition.

The high-level encoder architecture is similar to the Transformer model~\cite{vaswani2017attention}; however, we consider convolutions instead of self-attention. Our architecture is inspired by and quite related to the lightweight convolution~\cite{wu2019pay}. An important idea of that work and ours is the separation of the integration over time from the mixing over channels which improves both accuracy and efficiency. Other than the application to speech, some differences in our encoder architecture are (1) the time-depth separable convolution can be implemented with a simple 2D convolution and (2) our models do not use any normalization over the time dimension of the kernels.

Depth-wise separable convolutions have been used to improve the efficiency and accuracy of computer vision models~\cite{chollet2017xception, howard2017mobilenets}. 
The first layer of the TDS block can be seen as a grouped 1D convolution with $cw$ channels, a group size of $c$, and weights tied between groups. Grouped convolutions have also been used in computer vision to improve efficiency for e.g. model-parallel training~\cite{krizhevsky2012imagenet} and classification accuracy~\cite{xie2017aggregated}.

\section{Conclusion}

We have shown that a fully convolutional encoder and a simple decoder can give superior results to a strong RNN baseline while being an order of magnitude more efficient. Key to the success of the convolutional encoder is a time-depth separable block structure which allows the model to retain a large receptive field. We also show how to integrate a strong convolutional LM with a stable and scalable beam search procedure.

\section{Acknowledgements}

Thanks to Michael Auli, Abdelrahman Mohamed, Tatiana Likhomanenko and Gabriel Synnaeve for helpful conversations.

\bibliographystyle{IEEEtran}

\bibliography{refs}

\end{document}